\documentclass[conference]{IEEEtran}
\IEEEoverridecommandlockouts
\usepackage{needspace}
\usepackage{cite}
\usepackage{amsmath,amssymb,amsfonts}
\usepackage{booktabs}
\usepackage{algorithmic}
\usepackage{multirow}
\usepackage{graphicx}
\usepackage{hyperref} 
\usepackage{textcomp}
\usepackage{xcolor}
\def\BibTeX{{\rm B\kern-.05em{\sc i\kern-.025em b}\kern-.08em
    T\kern-.1667em\lower.7ex\hbox{E}\kern-.125emX}}
\begin{document}

\title{Single-Step Reconstruction-Free Anomaly Detection and Segmentation via Diffusion Models\\
}


\author{\IEEEauthorblockN{1\textsuperscript{st} Mehrdad Moradi}
\IEEEauthorblockA{\textit{H. Milton Stewart School of Industrial and Systems Engineering} \\
\textit{Georgia Institute of Technology}\\
Atlanta, Georgia, USA \\
mmoradi6@gatech.edu}
\and
\IEEEauthorblockN{2\textsuperscript{nd} Marco Grasso}
\IEEEauthorblockA{\textit{Department of Mechanical Engineering} \\
\textit{Polytechnic University of Milan}\\
Milan, Italy \\
marcoluigi.grasso@polimi.it}
\and
\IEEEauthorblockN{3\textsuperscript{rd} Bianca Maria Colosimo}
\IEEEauthorblockA{\textit{Department of Mechanical Engineering} \\
\textit{Polytechnic University of Milan}\\
Milan, Italy \\
biancamaria.colosimo@polimi.it}
\and
\IEEEauthorblockN{4\textsuperscript{th} Kamran Paynabar}
\IEEEauthorblockA{\textit{H. Milton Stewart School of Industrial and Systems Engineering} \\
\textit{Georgia Institute of Technology}\\
Atlanta, Georgia, USA \\
kamran.paynabar@isye.gatech.edu}
}

\maketitle
\begin{abstract}
Generative models have demonstrated significant success in anomaly detection and segmentation over the past decade. Recently, diffusion models have emerged as a strong alternative, outperforming previous approaches such as GANs and VAEs. In typical diffusion-based anomaly detection, a model is trained on normal data, and during inference, anomalous images are perturbed to a predefined intermediate step in the forward diffusion process. The corresponding normal image is then reconstructed through iterative reverse sampling.

However, reconstruction-based approaches present three major challenges: (1) the reconstruction process is computationally expensive due to multiple sampling steps, making real-time applications impractical; (2) for complex or subtle patterns, the reconstructed image may correspond to a different normal pattern rather than the original input; and (3) choosing an appropriate intermediate noise level is challenging because it is application-dependent and often assumes prior knowledge of anomalies, an assumption that does not hold in unsupervised settings.

We introduce \textit{\textbf{R}econstruction-free \textbf{A}nomaly \textbf{D}etection with \textbf{A}ttention-based diffusion models in Real-time (RADAR)}, which overcomes the limitations of reconstruction-based anomaly detection. Unlike current state-of-the-art (SOTA) methods that reconstruct the input image, RADAR directly produces anomaly maps from the diffusion model, improving both detection accuracy and computational efficiency. We evaluate RADAR on real-world 3D-printed material and the MVTec-AD dataset. Our approach surpasses state-of-the-art diffusion-based and statistical machine learning models across all key metrics, including accuracy, precision, recall, and F1 score. Specifically, RADAR improves F1 score by 7\% on MVTec-AD and 13\% on the 3D-printed material dataset compared to the next best model.

\textbf{Code:}
\href{https://github.com/mehrdadmoradi124/RADAR}{https://github.com/mehrdadmoradi124/RADAR}
\end{abstract}
\section{Introduction}
\begin{figure}[htbp]
    \centering
    \includegraphics[width=\linewidth]{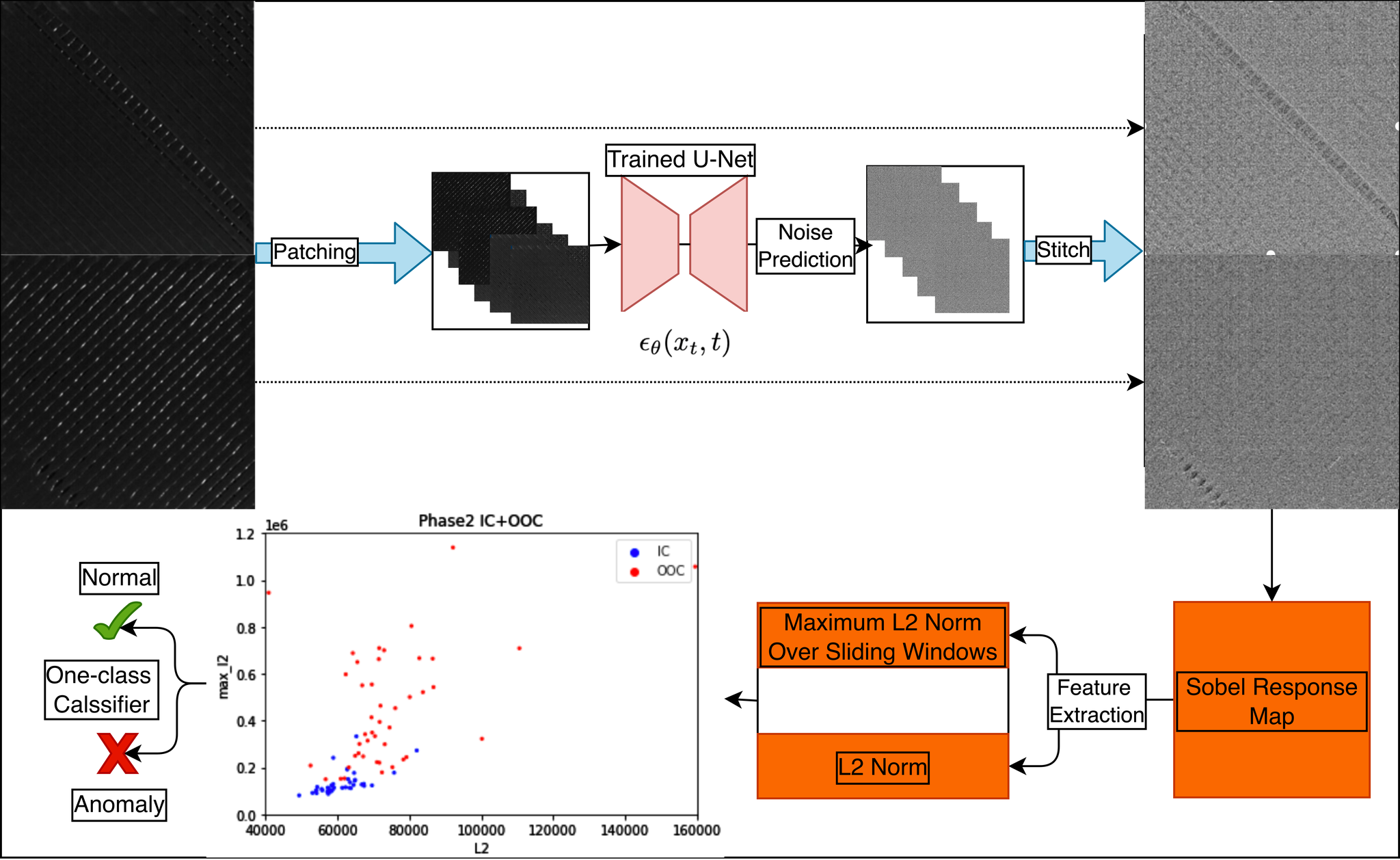}
    \caption{Overview of the RADAR anomaly detection framework. After training the diffusion model, each new image is divided into overlapping patches, and the predicted noise maps for these patches are combined to form the full-image noise map. Sobel edge detection is then applied to highlight the edges of anomalous regions. Features are extracted by computing the global L2 norm over the entire noise map and the maximum L2 norm within a sliding window for local anomaly characterization.}
    \label{fig:feature_ex}
\end{figure}
With significant advancements in sensing technology, the volume of image data collected for quality control has increased exponentially over the past decade. This rapid growth has created an increased demand for reliable, real-time algorithms capable of performing online anomaly detection and segmentation. Such capabilities are crucial across diverse industries, including stone countertop manufacturing\cite{liu2006estimation_stone}, ceramic capacitor chip production\cite{lin2007computer_ceramic}, softwood lumber grading\cite{bharati2003softwood_lumbar}, material microstructure monitoring\cite{liu2015random_material, torquato2002statistical_material}, fabric inspection\cite{bui2017monitoring_B&A}, and three-dimensional printing for additive manufacturing\cite{caltanissetta2024monitoring_bianca}.

One popular approach for anomaly detection involves low-rank matrix decomposition, where an input image is decomposed into a normal (low-rank) component and an anomalous (sparse) component. This paradigm is conceptually simple and has shown strong performance on structured patterns, with applications in PV panel anomaly detection\cite{wang2022online_pv,yang2025anomaly_pv}, periodic pattern data\cite{ye2025novel_psd}, and composite sheets\cite{yan2020akm2d}. Robust PCA (RPCA)\cite{candes2011robust_RPCA} achieves this decomposition by enforcing sparsity in the anomalous component using an L1 norm while minimizing the nuclear norm of the low-rank part. Smooth-Sparse Decomposition (SSD)\cite{yan2017anomaly_SSD} extends this idea by modeling both components using B-spline bases and assuming smoothness in the normal part. Despite their success, these methods rely heavily on the assumption that normal data are approximately low-rank and anomalies are sparse. When these assumptions are violated—such as in datasets with complex nonlinear textures—their performance deteriorates.

To handle nonlinear and complex patterns, various machine learning approaches have been proposed. In supervised settings, methods such as\cite{wang2020cnn,inproceedings_cnn_svm,jin2019autonomous_cnn} employ deep convolutional neural networks (CNNs) to detect anomalies. Unlike supervised methods, unsupervised anomaly detection has no access to anomalous images or any form of anomaly annotations (pixel-level or image-level), relying solely on normal data to learn the underlying data distribution. In this context, \cite{bui2017monitoring_B&A} trained a model to predict each pixel’s value based on its surrounding pixels within a local window, using only normal data during training. During inference, the Anderson-Darling statistic was computed to evaluate the deviation of residuals from the empirical normal distribution. Caltanissetta et al. \cite{caltanissetta2024monitoring_bianca} extended this approach by using a larger prediction window and introducing more robust statistical measures for anomaly detection. While these models achieve strong performance when trained on a single dominant pattern, they require separate models for each distinct pattern in the data. Consequently, their performance degrades significantly when multiple patterns coexist in the training set.

To model complex or multi-modal training distributions, generative models are widely employed to learn the underlying data distribution. These models have achieved remarkable success in learning high-dimensional data distributions and generating realistic images. Among state-of-the-art generative models, diffusion models\cite{ho2020ddpm} have outperformed GANs\cite{goodfellow2014generative} and VAEs\cite{kingma2013auto} in terms of image quality, mode coverage, and training stability\cite{dhariwal2021diffusion_beat, wyatt_anoddpm_2022}. Typically, diffusion models are trained exclusively on normal data. During inference, anomaly detection is performed by computing the residual between the input image and its reconstruction. Although this approach has achieved significant success, it struggles when anomalies are subtle—i.e., when intensity differences between normal and anomalous regions are minimal. It also fails to generalize well when the training dataset is small. Another limitation is the need for multiple sampling steps during reconstruction, which hinders real-time implementation.

To overcome the limitations of reconstruction-based generative models, we propose Reconstruction-free Anomaly Detection with Attention-based Diffusion Models in Real-time (RADAR). Unlike conventional diffusion methods that require iterative sampling to reconstruct the input image, RADAR directly produces anomaly maps in a single forward step, enabling real-time implementation. By leveraging a patch-based learning strategy, RADAR not only augments the training data but also improves generalization while reducing the computational overhead of the attention mechanism. Moreover, our approach avoids the corruption of input images during inference, which has a twofold advantage: (1) normal patterns remain intact, allowing the anomaly map to accurately reflect the true normal features rather than a reconstructed approximation, and (2) subtle anomalies are preserved instead of being degraded by forward diffusion steps, enabling more precise localization of anomalous regions.

The structure of this paper is as follows. We review the related literature on image-based anomaly detection for patterned data in Sec.~\ref{related}, present the problem formulation and our proposed methodology in Sec.~\ref{method}, evaluate the effectiveness of RADAR through extensive experiments in Sec.~\ref{exp}, and conclude the paper in Sec.~\ref{conc}. Our key contributions are summarized below:
\begin{itemize}
    \item We introduce a new reconstruction-free anomaly detection paradigm. Instead of generating an anomaly-free image, RADAR directly computes anomaly maps using the diffusion model. We justify this approach using the forward and backward diffusion Markov chains.
    \item By avoiding iterative sampling and applying the diffusion model only once, our method achieves real-time anomaly detection while improving the performance.
    \item We leverage patch-based training, reducing GPU memory usage by 400× while improving performance in low-data settings. Although this increases training time (about 25× longer), it can be mitigated by adjusting training epochs.
\end{itemize}
\section{Related Work}
\label{related}
\subsection{Statistical and Deep Learning Methods}
A widely used approach for image anomaly detection is low-rank decomposition, where an image is modeled as the sum of a low-rank component (normal patterns) and a sparse component (anomalies). Robust PCA (RPCA)\cite{candes2011robust_RPCA} enforces low-rankness using nuclear norm minimization and sparsity using an $\ell_1$-norm penalty, effectively separating structured background from sparse anomalies. Smooth-Sparse Decomposition (SSD)\cite{yan2017anomaly_SSD} extends RPCA by incorporating B-spline bases to enforce smoothness in the low-rank component. Periodic-Sparse Decomposition (PSD)\cite{ye2025novel_psd} addresses images with repetitive periodic patterns by formulating periodicity as an optimization constraint while keeping the anomaly component sparse. Similarly, \cite{yang2025anomaly_pv} demonstrated the effectiveness of RPCA for anomaly detection in PV panel images by leveraging smoothness across periodic signals.
However, the low-rank assumption is often unrealistic for real-world data where normal patterns can be highly textured, nonlinear, or have complex structures that are not well-approximated by a low-rank subspace. Moreover, anomalies themselves are not always sparse or well-isolated, which further limits the effectiveness of these methods. To address these limitations, \cite{mou2024paedid} relaxed the low-rank assumption by learning a deep image prior using an autoencoder trained on normal images, and then decomposed anomaly images into normal and sparse anomaly components.

Convolutional neural networks (CNNs) have been frequently used for anomaly detection and segmentation\cite{wang2020cnn,inproceedings_cnn_svm,jin2019autonomous_cnn}. More recently, researchers have explored the use of pre-trained models to extract and compare image embeddings. For example, \cite{wan2021industrial_pretrained,liu2023unsupervised_pretrained} applied Gaussian mixture models to the embeddings derived from pre-trained networks and detected anomalies by measuring the Mahalanobis distance between the test image embedding and the distribution of normal embeddings. Similarly, \cite{wan2022unsupervised_pretrained} trained a neural network to learn mappings between embeddings from two different pre-trained models and identified anomalies using the mean squared error between predicted and actual embeddings.

In contrast, our method does not rely on low-rank assumptions, pre-trained models, or prior knowledge of anomalies. We operate in a fully unsupervised setting with only normal data available during training. Our diffusion-based approach directly learns the distribution of normal data and produces pixel-level anomaly maps that effectively distinguish in-distribution from out-of-distribution regions.
\subsection{Diffusion Models for Anomaly Detection}
Diffusion models have demonstrated strong performance in image synthesis\cite{dhariwal2021diffusion_beat}. In the context of anomaly detection, a diffusion model is trained exclusively on normal data and then used to reconstruct anomalous samples by generating their corresponding “normal” counterparts. If the training data are contaminated, RDDPM\cite{moradi2025rddpm} uses the Huber loss to achieve robustness to outliers. These models can be viewed as approximating a stochastic differential equation (SDE) in both the forward and reverse diffusion processes\cite{song2020score_sde}, making the sampling path inherently stochastic. Guidance during the reverse sampling process is often necessary to ensure that the reconstructed image preserves the structural features of the input anomaly image. A key limitation of diffusion models is their high computational cost, as achieving high-quality reconstructions typically requires hundreds of reverse steps\cite{ho2020ddpm}. To address this, latent diffusion models\cite{rombach2022high_stable} operate in the lower-dimensional embedding space of an encoder, reducing computational complexity while maintaining fidelity.

Backward diffusion can be guided by incorporating additional information into the sampling process. Classifier guidance introduces the gradient of a classifier loss during reverse sampling, encouraging the generation of samples that align with specific target classes or features\cite{ho2022classifier_free}. Classifier-free guidance\cite{ho2022classifier_free} achieves a similar effect without explicitly training a classifier, by conditioning the diffusion model on auxiliary inputs (e.g., image embeddings or prompts) through mechanisms such as cross-attention. Finally, implicit guidance\cite{wyatt_anoddpm_2022} initializes the reverse process with an intermediate noised version of the input image rather than pure Gaussian noise, ensuring that the reconstruction retains the structure of the original image while producing its anomaly-free version.

A key challenge of reconstruction-based diffusion models for anomaly detection is preserving structural fidelity to the input image while removing anomalies. To address this, \cite{zhang2023unsupervised_diffad} proposed a latent diffusion model conditioned on the noisy input image to better retain its content. Other approaches incorporate semantic-guided networks to ensure that reconstructed images belong to the correct category in multi-class anomaly detection scenarios\cite{zhang2025dzad,he2024diffusion_diad}. In\cite{zhang2025diffusionad}, the number of reverse sampling steps was significantly reduced by combining classifier-like guidance with implicit guidance. Specifically, they estimate $x_0$ directly from intermediate noisy states by conditioning the predicted noise on higher-noise reconstructions via an $\ell_2$ norm loss. Meanwhile, \cite{tebbe2024dynamic} proposed dynamic noise addition, where the forward diffusion step $t$ is chosen adaptively based on the feature similarity between the input image and normal samples from a pre-trained model, using larger noise levels for more pronounced anomalies.

Several studies have introduced auxiliary networks to enhance the performance of diffusion models for reconstruction or anomaly  segmentation. For instance, \cite{yan2023feature_video} utilized two diffusion models: the first model, conditioned on temporal video frames, captures motion features, while the second, conditioned on the reconstruction from the first, focuses on spatial features for video anomaly detection. Similarly, \cite{pintilie2023time_AEDIFF} integrated a diffusion model with an encoder-decoder architecture, training both networks end-to-end to improve reconstruction fidelity. In~\cite{flaborea2023multimodal_AE_condition}, an autoencoder was combined with a diffusion model to inject temporal features into the diffusion process. Finally, other works have added segmentation sub-networks to diffusion models to directly improve pixel-level anomaly localization\cite{zhang2023unsupervised_diffad,zhang2025diffusionad}.

Despite recent advancements, reconstruction-based approaches still struggle to capture subtle anomaly patterns and perform poorly in low-data regimes. Additionally, their high computational cost due to iterative sampling makes them impractical for real-time industrial applications. In contrast, our method directly generates anomaly maps in a single forward pass of the diffusion model, eliminating the need for reconstruction. As a result, the generated anomaly maps are more precise, since our approach avoids the error accumulation seen in multi-step reconstruction pipelines.
\section{Methodology}
\label{method}
\subsection{Problem Formulation and Overview}
Given a set of normal training images $\mathcal{D} = \{ x_i \}_{i=1}^N$, where N is the total number of training samples, our objective is to learn a model f that can determine whether a new test image $\hat{x}$ is anomalous (label =1) or normal (label =0) based on its deviation from the distribution of normal data. Importantly, this is achieved without access to pixel-level anomaly annotations, which are expensive to collect, or any anomalous images during training.

To construct f, we train a diffusion U-Net $\epsilon_\theta$ exclusively on normal data. During inference, we compute a compact feature representation by applying a Sobel edge detector and the L2 norm to the outputs of $\epsilon_\theta$ on noisy versions of the input image. The Sobel operator highlights local structural changes (e.g., edges), while the L2 norm aggregates these variations into a feature vector. An overview of our method is shown in Fig.~\ref{fig:feature_ex}, and each step is detailed in the following subsections.
\subsection{Diffusion Model Background}
\begin{figure}[htbp]
    \centering
    \includegraphics[width=\linewidth]{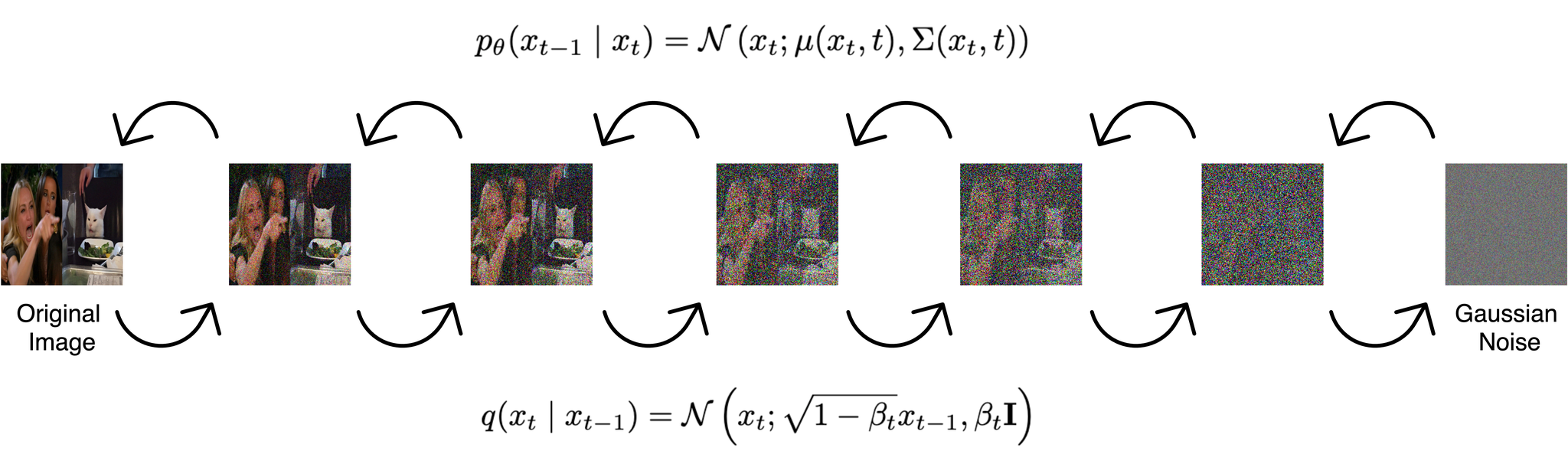}
    \caption{Forward and backward diffusion processes. The forward process gradually adds noise to an image until it becomes random noise, while the backward process uses a trained neural network to progressively remove the noise and recover the original image structure.}
    \label{fig:forwardbackward}
\end{figure}

\begin{figure}[htbp]
    \centering
    \includegraphics[width=\linewidth]{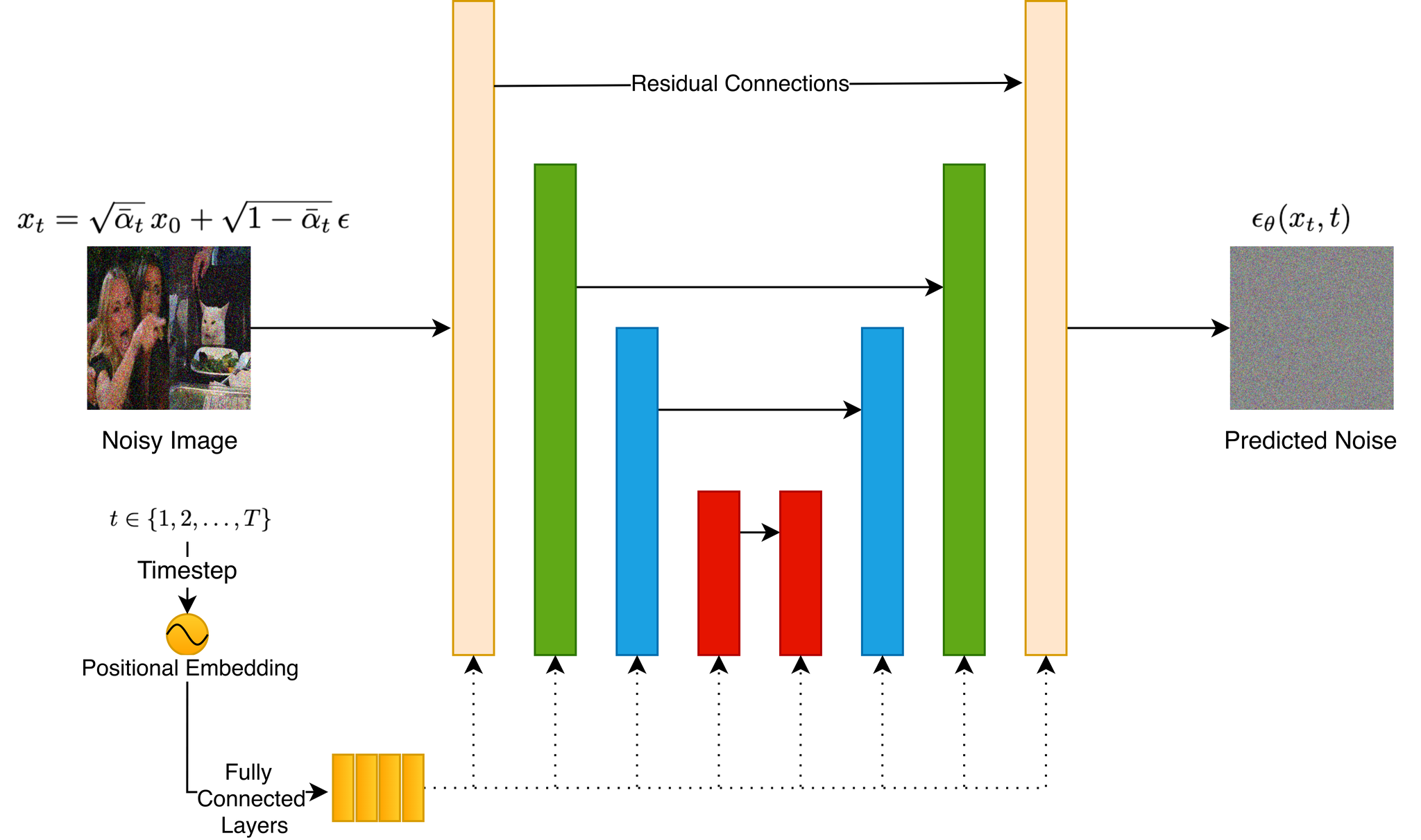}
    \caption{DDPM as a noise prediction model. A Gaussian noise and time step are sampled to perturb the image via the forward diffusion process. The noisy image is then passed through a U-Net, which predicts the added noise. The time step is encoded into positional embeddings, processed by a fully connected network, and injected into all U-Net layers.}
    \label{fig:noise_pred}
\end{figure}

Denoising Diffusion Probabilistic Models (DDPMs)\cite{ho2020ddpm} are a class of generative models that learn data distributions by gradually adding Gaussian noise in a forward process and then reversing this process. The forward diffusion process is modeled as a Markov chain that incrementally perturbs the data:
\begin{equation}
\label{eq:forward_ddpm}
q(\mathbf{x}t | \mathbf{x}{t-1}) = \mathcal{N}(\mathbf{x}t; \sqrt{1 - \beta_t} , \mathbf{x}{t-1}, \beta_t \mathbf{I}),
\end{equation}
where $t \in \{1, \ldots, T\}$ is the time step, $\mathbf{x}_t$ is the noisy sample at step t, and $\beta_t$ is the predefined noise schedule.
Fig.~\ref{fig:forwardbackward} provides an overview of this forward and backward process.
To recover the original data, the reverse process is approximated as a Gaussian distribution with a neural network predicting the mean:
\begin{equation}
p_\theta(\mathbf{x}_{t-1} | \mathbf{x}_t) = \mathcal{N} \left( \mathbf{x}_{t-1}; \mu_{\theta}(\mathbf{x}_t, t), \sigma_t^2 \mathbf{I} \right),
\end{equation}
where
$\mu_{\theta}(\mathbf{x}_t,t)=
\frac{1}{\sqrt{\alpha_t}} \left( \mathbf{x}_t - \frac{\beta_t}{\sqrt{1 - \bar{\alpha}t}} \, \epsilon_{\theta}(\mathbf{x}_t, t) \right), \alpha_t = 1 - \beta_t, \bar{\alpha}_t = \prod_{i=1}^t \alpha_i.$
The model parameters $\theta$ are optimized by minimizing the following simplified objective:
\begin{equation}
\label{eq:Lsimple}
L_{\text{simple}}(\theta) := \mathbb{E}_{t, \mathbf{x}0, \boldsymbol{\epsilon}} \left[
\left| \boldsymbol{\epsilon} - \epsilon_\theta \left(
\sqrt{\bar{\alpha}_t} \mathbf{x}_0 + \sqrt{1 - \bar{\alpha}t} \boldsymbol{\epsilon}, t
\right) \right|^2
\right],
\end{equation}
where $\boldsymbol{\epsilon} \sim \mathcal{N}(0, \mathbf{I})$.
This loss effectively trains $\epsilon\theta$ to predict the noise added at each step, commonly implemented with a U-Net architecture\cite{ronneberger_u-net_2015}, as illustrated in Fig.~\ref{fig:noise_pred}.
\vspace{-2mm}
\subsection{Single-step Anomaly Map Generation}
Over the past decade, generative models for anomaly detection have primarily been trained on normal data to learn the underlying distribution of normal patterns. During inference, these models attempt to reconstruct the input by generating its closest normal counterpart.

When using diffusion models for reconstruction, the anomalous image is first corrupted by adding noise up to a certain level. This noise injection serves two purposes: (1) it disrupts anomaly pixels so they do not influence the reverse generation process, and (2) it preserves normal pixels sufficiently to act as implicit guidance during backward diffusion. However, achieving this balance is challenging—especially when anomaly pixels exhibit smaller intensity values than normal patterns—because no single noise step can both erase anomalies and retain fine normal features.
Furthermore, reconstruction-based approaches are computationally expensive due to the need for multiple iterative sampling steps in reverse diffusion, sometimes up to 500 steps. In low-data engineering applications, they often overfit and fail to accurately reconstruct complex patterns.

\begin{figure}[htbp]
    \centering
    \includegraphics[width=\linewidth]{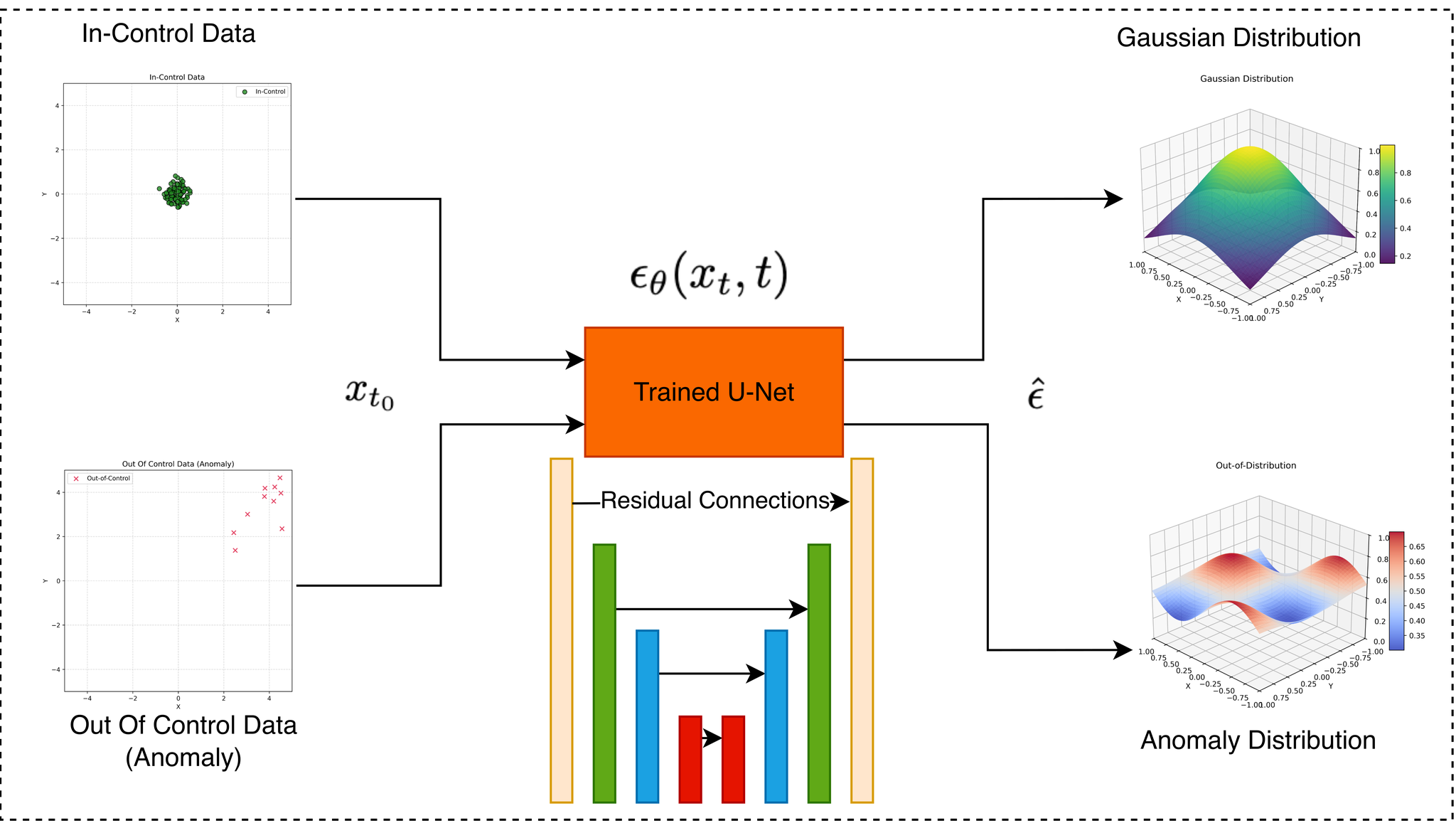}
    \caption{Single-step anomaly map generation. For normal (in-control) data, the trained diffusion model predicts noise that follows a Gaussian distribution due to the Gaussian nature of the forward and backward diffusion processes. In contrast, anomalous patches produce non-Gaussian noise patterns, allowing the model to identify anomalies.}
    \label{fig:anomaly_map}
\end{figure}

To address these limitations, we propose a fundamentally different paradigm: instead of reconstructing the input image, we directly leverage the trained U-Net to produce an anomaly map. For normal training data, the predicted noise approximates a Gaussian distribution because the forward diffusion process injects standard Gaussian noise, and the L2 loss in Eq.~\ref{eq:Lsimple} encourages the model to match this distribution. In contrast, anomalous inputs lead to noise predictions that deviate from this Gaussian behavior. To preserve the fine details of the input, we apply only a single forward diffusion step, which minimally affects the original image. The predicted noise at this step is then used as the anomaly signal, as illustrated in Fig.~\ref{fig:anomaly_map}.

To handle low-data scenarios in engineering applications such as additive manufacturing, we adopt a patch-based training strategy by extracting small patches from the training images. This increases the training dataset size, reduces overfitting, and significantly lowers computational cost. For example, dividing a 500×500 image into 25×25 patches reduces GPU memory usage to only 0.25\% (1/400) of what would be required for full-image training.

If 100×100 overlapping patches are extracted per image, the number of training samples scales by 100×100 = 10,000. Considering that each 25×25 patch is 1/400th of the size of the original image, the total training time effectively increases by only:
$\frac{100 \cdot 100}{20 \cdot 20} = 25 \text{ times}$.
This increase can be offset by reducing the number of epochs due to the enlarged dataset. Moreover, patch-based learning improves generalization by exposing the model to diverse localized features.

An example output of the predicted noise is shown in Fig.~\ref{fig:feature_ex}. Due to the distribution shift, the predicted noise behaves very differently in anomalous pixels but remains consistent in normal regions, regardless of the underlying normal pattern. In the next step, features are extracted from this noise map.
\subsection{Feature Extraction}
The predicted noise map exhibits distinctive textural patterns in anomalous regions compared to normal areas due to the shift in the distribution. We leverage image processing techniques to extract features that are subsequently used for anomaly detection and segmentation. As a first step, we apply a Gaussian or uniform blur to smooth the entire map, making the normal regions more uniform and enhancing the contrast between normal and anomalous textures. 

We apply Sobel edge detection to highlight the boundaries of anomalous regions. To generate a low-dimensional feature vector, we compute the overall L2 norm of the Sobel map to represent the global anomaly feature. For local feature extraction, a sliding window is applied, and the L2 norm is computed within each window. The maximum of these local norms is selected as the local anomaly feature. Together, these global and local features form a compact 2D representation of the image, which is subsequently used for unsupervised one-class classification. 

Fig.~\ref{fig:feature_ex} visualizes the two-dimensional feature space for both in-control and out-of-control data from Phase 2 of our first case study on 3D printed material images. The features reveal a clear separation between the two classes, paving the way for effective binary classification in the next step.
\vspace{-2mm}
\subsection{Classification}
To classify whether an image is anomalous, we train a one-class classifier on the feature vectors extracted from normal training data. Any novelty detection algorithm could be applied in this step, but among classical approaches, Isolation Forest (IF)\cite{iso_forest} provides notable advantages over methods such as Local Outlier Factor (LOF)\cite{breunig2000lof} and One-Class SVM (OC-SVM)\cite{scholkopf1999svm}. Unlike LOF and OC-SVM, which rely on density estimation or kernel-based decision boundaries, IF isolates anomalies by recursively partitioning the feature space, leading to $O(n \log n)$ scalability. In contrast, OC-SVM has a higher computational complexity of $O(n^3)$\cite{kang2019approximate_svm_complexity}, while LOF scales at $O(n^2)$\cite{alghushairy2020lof_complexity}, making them less suitable for large datasets. 

In our experiments, we set the contamination level to 0.05 and used 100 estimators. If validation data is available, the method can be applied in a semi-supervised manner by fine-tuning the parameters using the validation set before evaluating on the test data.
\subsection{Pixel Level Anomaly Map}
The single-step nature of our algorithm enables real-time implementation across various engineering applications. Once a sample is identified as anomalous, pixel-level localization is achieved by applying Sobel edge detection to a blurred version of the image, where blurring reduces noise and emphasizes anomaly boundaries. The Sobel output highlights the pixels corresponding to anomalous regions. Representative qualitative examples are shown in Figs.~\ref{fig:print_qualitative} and~\ref{fig:tile_qualitative}.
\section{Experiments and Results}
\label{exp}
\subsection{Experimental Setup}
\subsubsection{Datasets}
We use the dataset introduced in \cite{caltanissetta2024monitoring_bianca}, which contains 3D printed material images from the Fused Filament Fabrication (FFF) process with two distinct deposition strategies (45° and 135°). These deposition angles are chosen to represent different structural patterns in the printed material. Examples of defective samples are shown in the left column of Fig.~\ref{fig:print_qualitative}. Additionally, we use 107 images from the tile category of the MVTec-AD dataset\cite{bergmann_mvtec_2019}, which contains highly stochastic surface patterns. These textures complement the periodic patterns observed in the 3D Print dataset. 

\textit{3D Prints} (FFF) contains two deposition patterns, \(45^\circ\) and \(135^\circ\): 
training has 81 normal images (40 @ \(135^\circ\), 41 @ \(45^\circ\);
testing has 84 images—40 normal (19 @ \(135^\circ\), 21 @ \(45^\circ\)) and 44 anomalous (22 @ \(135^\circ\), 22 @ \(45^\circ\)). 
\textit{MVTec-AD Tile} has 230 normal training images and 107 test images comprising 33 normal and 74 anomalous across five defect types: crack (15), glue strip (16), gray stroke (14), oil (16), and rough (13).
\subsubsection{Implementation Details}
We use grayscale images resized to 100 × 100 pixels for training. Each image is divided into 73 × 73 overlapping patches, where each patch has a size of 28 × 28 pixels. This results in 431,649 training samples for the 3D Print dataset and 1,225,670 samples for the Tile dataset. Due to the large number of training samples in both datasets, we use three training epochs for the 3D Print dataset and six epochs for the Tile dataset. The Tile dataset requires more training epochs because its normal patterns are highly stochastic, whereas the 3D Print dataset exhibits regular periodic patterns.

\begin{figure}[htbp]
    \centering
    \includegraphics[width=\linewidth]{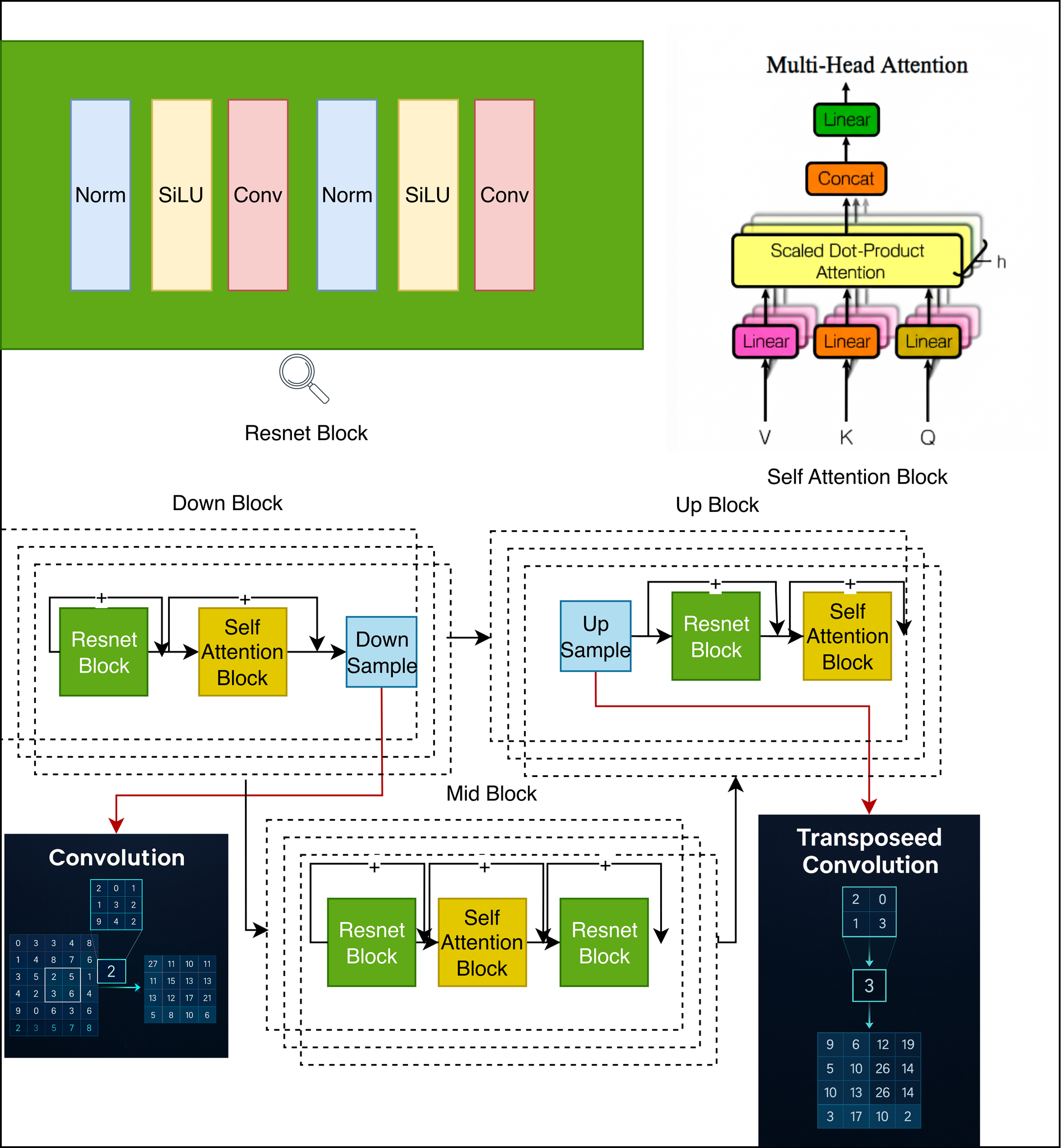}
    \caption{U-Net architecture. The network consists of a downsampling block, a middle block, and an upsampling block. Each block integrates attention modules, convolutional layers, and residual connections, which are represented by arrows with a plus sign. Convolution, transposed convolution and multi-head attention pictures were generated by GPT-4o \cite{openai_gpt4o_systemcard_2024}.}
    \label{fig:arch}
\end{figure}

The U-Net architecture we use closely follows the design of the stable diffusion model\cite{rombach2022high_stable}. As shown in Fig.~\ref{fig:arch}, it consists of a downsampling block, an upsampling block, and a mid block, all connected through residual and skip connections. Each block is composed of ResNet blocks, self-attention modules, and either downsampling or upsampling layers. For downsampling, a 2D convolution with a kernel size of 4, stride of 2, and padding of 1 is applied to halve the spatial resolution. Conversely, for upsampling, we use a transposed 2D convolution (ConvTranspose2d) with the same kernel size, stride, and padding to double the resolution. Each ResNet block includes a normalization layer, SiLU activation functions, and convolutional operations. The self-attention module leverages learnable keys, queries, and values to compute attention weights across all spatial positions, enabling the model to capture long-range dependencies. This mechanism allows the U-Net to focus on the most informative regions of the feature maps, improving its ability to detect subtle anomalies that may not be captured by purely convolutional operations.

For the diffusion model, we adopt the Denoising Diffusion Probabilistic Model (DDPM) with 1000 forward diffusion steps, following the setup in the original work\cite{ho2020ddpm}. The noise schedule linearly increases from 0.0001 to 0.02. Training is done with a batch size of 64 and a learning rate of 0.0001. All models were trained using an NVIDIA RTX6000 GPU. 
\subsubsection{Benchmarks}
We compare our approach against three state-of-the-art diffusion-based anomaly detection models: AnoDDPM\cite{wyatt_anoddpm_2022}, DiffusionAD\cite{zhang2025diffusionad}, and the original DDPM\cite{ho2020ddpm}. These reconstruction-based methods have demonstrated state of the art performance in anomaly detection and segmentation tasks. Additionally, we benchmark our method against two statistical machine learning approaches—B\&A\cite{bui2017monitoring_B&A} and L\&F\cite{caltanissetta2024monitoring_bianca}, which have achieved state-of-the-art performance for unsupervised anomaly detection in stochastic and periodic pattern images.

For reconstruction-based anomaly detection methods, the residual between the input and reconstructed image is first computed. Otsu’s thresholding\cite{4310076_otsu} is then applied to the residual map to obtain a binary anomaly map. Finally, the L1 norm (sum of pixel values) of the binary map is used as the anomaly feature for detection.
\subsubsection{Evaluation Metrics}
The performance of our model is evaluated with accuracy, precision, recall, and F1 score. Accuracy shows the overall correctness of the model’s predictions but does not differentiate between the performance on positive and negative classes. Precision measures the proportion of correctly detected anomalies among all predicted anomalies. Recall, on the contrary, measures the proportion of actual anomalies correctly detected by the model. As the harmonic mean of precision and recall, the F1 score provides a balanced metric that reflects both false positives and false negatives.
\subsection{Anomaly Detection Results}
Tab.~\ref{tab:3dprints_tile} presents the performance of RADAR compared to competing methods. On the 3D printed material dataset, RADAR consistently outperforms all other models across all metrics. Notably, reconstruction-based diffusion models perform poorly, highlighting their limitations when anomaly patterns closely resemble normal patterns in pixel intensity. The best-performing diffusion-based method, DiffusionAD, achieves an F1 score of only 0.27—well below the 0.5 F1 score expected from a random classifier in a balanced binary setting. In fact, AnoDDPM and DiffusionAD, despite incorporating simplex noise and norm-guided reconstruction respectively, underperform compared to the baseline DDPM. In contrast, the statistical machine learning models L\&F and B\&A show strong performance, with B\&A matching RADAR’s precision (0.77) and achieving the second-highest F1 score of 0.72, trailing RADAR’s 0.85.

\begin{figure}[ht]
    \centering
    \begin{minipage}{0.48\textwidth}
        \includegraphics[height=3cm,width=\linewidth]{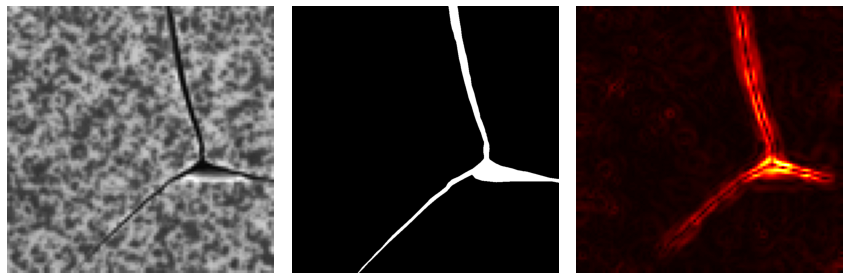} \\
        \includegraphics[height=3cm,width=\linewidth]{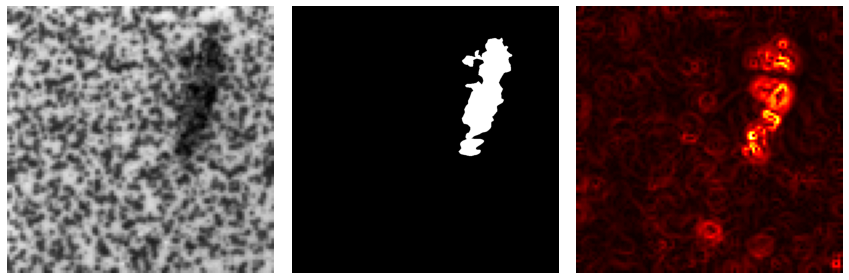}
    \end{minipage}
    \caption{Qualitative results on the Tile dataset. From left to right: defective images with crack and gray-stroke anomalies, the ground truth masks, and the pixel-level anomaly maps predicted by RADAR.}
    \label{fig:tile_qualitative}
\end{figure}

Compared to the 3D Print dataset, the metrics on the Tile dataset are overall lower, indicating that its stochastic patterns pose a greater challenge for our method. Interestingly, unlike with 3D Prints, reconstruction-based diffusion models outperform the statistical approaches on this dataset. AnoDDPM achieves the second-best F1 score of 0.60 (compared to RADAR’s 0.67), suggesting that the simplex noise mechanism in AnoDDPM is particularly effective for stochastic patterns. As illustrated in Fig.~\ref{fig:tile_qualitative}, anomalies in the Tile dataset exhibit clear pixel-intensity differences from normal patterns, which likely benefits diffusion-based methods. However, DiffusionAD performs poorly, implying that its norm-guided, two-step reconstruction is unsuitable for complex stochastic textures. Another notable observation is the poor performance of L\&F compared to B\&A, despite their conceptual similarities. This difference may stem from how the anomaly statistics are computed in L\&F. For this dataset, RADAR achieves the highest F1 score and accuracy overall. While L\&F attains the highest precision (0.95), its recall is only 0.07, indicating zero false positives but many missed anomalies in the form of false negatives. Conversely, DDPM and AnoDDPM achieve the highest recall (0.57), slightly outperforming RADAR’s recall of 0.51.
\renewcommand{\arraystretch}{2.5}
\begin{table}[ht]
\centering
\resizebox{0.48\textwidth}{!}{\fontsize{25pt}{16pt}\selectfont
\begin{tabular}{lcccccc}
\toprule
\textbf{Method} & \textbf{B\&A} & \textbf{L\&F} & \textbf{DiffusionAD} & \textbf{AnoDDPM} &\textbf{DDPM} & \textbf{RADAR (Ours)} \\
\midrule
\multicolumn{6}{c}{\textbf{3D Prints}} \\
\midrule
Accuracy$\uparrow$   & 0.73 & 0.67 &      0.42 &       0.46&  0.5& \textbf{0.82} \\
Precision$\uparrow$   & \textbf{0.77} & 0.70 &      0.39 &      0.45 & 0.6& \textbf{0.77} \\
Recall$\uparrow $     & 0.68 & 0.64 &      0.20 &      0.11 & 0.14& \textbf{0.93} \\
F1 Score$\uparrow$    & 0.72 & 0.67 &      0.27 &     0.18  & 0.22& \textbf{0.85} \\
\midrule
\multicolumn{6}{c}{\textbf{Tile}} \\
\midrule
Accuracy$\uparrow$    & 0.51 & 0.36 &     0.35  &     0.47  & 0.43& \textbf{0.64} \\
Precision$\uparrow$   & 0.87 & \textbf{1.00} &     0.58  &     0.63  & 0.59& 0.95 \\
Recall$\uparrow $     & 0.35 & 0.07 &      0.20 &    \textbf{0.57}   & \textbf{0.57}& 0.51 \\
F1 Score$\uparrow$   & 0.50 & 0.13 &     0.30  &      0.60 & 0.58& \textbf{0.67} \\
\bottomrule
\end{tabular}}
\vspace{0.5em}
\caption{Comparison of methods across Accuracy, Precision, Recall, and F1 Score for 3D Prints and Tile datasets.}
\label{tab:3dprints_tile}
\end{table}
\begin{figure}[ht]
    \centering
    \begin{minipage}{0.4\textwidth}
        \includegraphics[height=3cm,width=\linewidth]{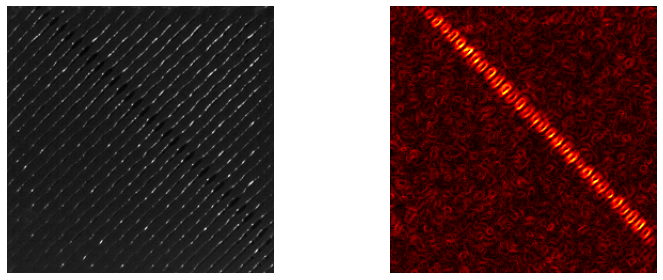} \\
        \includegraphics[height=3cm,width=\linewidth]{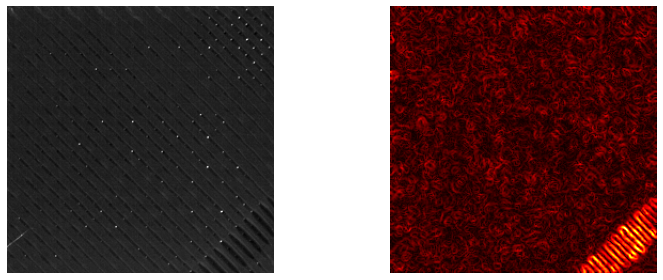}
    \end{minipage}
    \caption{Qualitative results on the 3D Print dataset. The left column displays defective images with 45° and 135° deposition patterns, while the right column shows the pixel-level anomaly heatmaps predicted by RADAR.}
    \label{fig:print_qualitative}
\end{figure}
\vspace{-2mm}
\subsection{Pixel Level Anomaly Segmentation}
The qualitative results shown in Figs.~\ref{fig:print_qualitative} and~\ref{fig:tile_qualitative} present the segmentation masks of sample images identified as anomalous by the algorithm. In Fig.~\ref{fig:print_qualitative}, both 45° and 135° deposition patterns with various types of anomalies are visualized, where the anomaly heatmaps on the right accurately localize the anomalous pixels. Fig.~\ref{fig:tile_qualitative} illustrates anomalies such as cracks and gray strokes on tile images, along with their corresponding heatmaps.
\subsection{Sensitivity Analysis}
\begin{figure}[htbp]
    \centering
    \includegraphics[width=\linewidth]{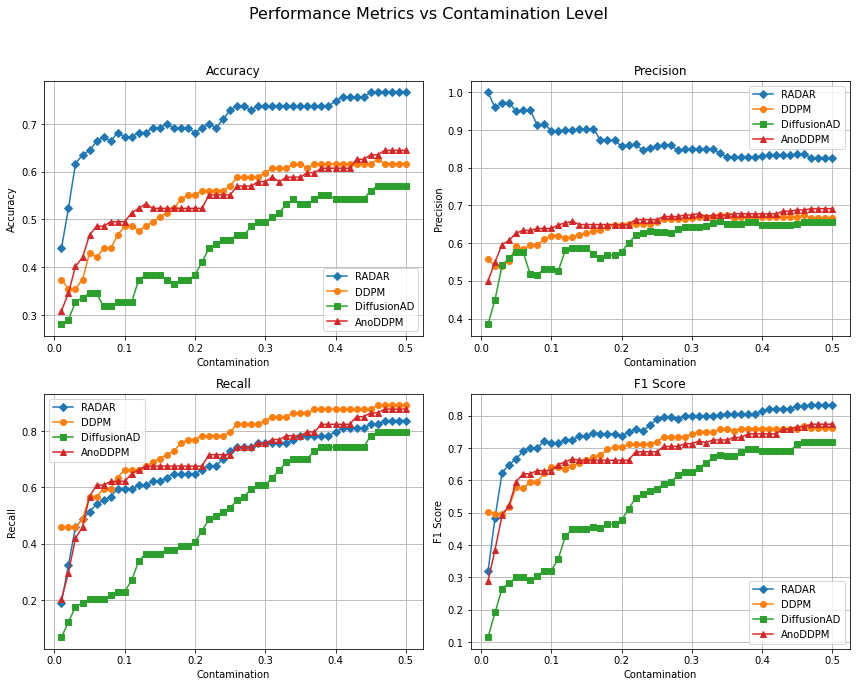}
    \caption{Contamination Level Sensitivity Analysis. The top row (left to right) shows accuracy and precision, while the bottom row (left to right) shows recall and F1 score. RADAR, DDPM, DiffusionAD, and AnoDDPM are represented by blue, orange, green, and red curves, respectively. The x-axis denotes the contamination level, and the y-axis indicates the corresponding metric.}
    \label{fig:sensitivity}
\end{figure}

As discussed earlier, the contamination level must be set when using the Isolation Forest algorithm for one-class classification. In our experiments, we fixed this parameter at 0.05. To evaluate its impact, we varied the contamination level from 0 to 0.5 and plotted all metrics for each diffusion-based model on the Tile dataset. As illustrated in Fig.~\ref{fig:sensitivity}, our model consistently outperforms all other methods in terms of accuracy, precision, and F1 score, though it slightly underperforms in recall compared to DDPM and AnoDDPM.
\section{Conclusion}
\label{conc}
Reconstruction-based anomaly detection methods have long faced challenges related to image fidelity and computational cost. In this paper, we introduced RADAR, a reconstruction-free anomaly detection and segmentation diffusion model that operates in a single step. RADAR leverages the predicted noise from the diffusion process to identify out-of-distribution data, and image processing techniques are employed to extract low-dimensional representations for anomaly detection. Its patch-based learning strategy makes it applicable for small training data. Our method demonstrated strong performance on both periodic and stochastic patterned datasets where reconstruction-based or statistical models struggle. 

Future work will focus on extending RADAR to other data modalities, including non-stationary time series and point clouds.
\section{Acknowledgment}
\label{ack}
Prof. Bianca Maria Colosimo’s research was partially funded by MICS (Made in Italy – Circular and Sustainable) Extended Partnership received funding from the European Union Next-GenerationEU (PIANO NAZIONALE DI RIPRESA E RESILIENZA (PNRR) – MISSIONE 4 COMPONENTE 2, INVESTIMENTO 1.3 – D.D. 1551.11-10-2022, PE00000004).
\bibliographystyle{IEEEtran}
\bibliography{references}
\end{document}